\documentclass[letterpaper]{article} 
\usepackage[preprint]{aaai2027}  
\usepackage[hyphens]{url}  
\usepackage{graphicx} 
\usepackage{natbib}  
\usepackage{caption} 
\usepackage{booktabs}
\usepackage{newfloat}
\DeclareFloatingEnvironment[fileext=loa,name=Algorithm,placement=t]{algorithm}
\usepackage{amsmath,amssymb}
\newcommand{\psopsd}{\textsc{PS-OPSD}}
\newcommand{\opsd}{\textsc{OPSD}}
\newcommand{\psguide}{g_{\mathrm{PS}}}
\newcommand{\problemspace}{\mathfrak{P}}
\newcommand{\student}{p_{\theta}}
\newcommand{\teacher}{p_{T}}

\title{Is More Privileged Information Better? From Solution Traces to Problem-Solving Structure in Self-Distilled Reasoning}
\author{
    Xuyang Zhao\textsuperscript{\rm 1},
    Liting Zhang\textsuperscript{\rm 1},
    Zichen Xu\textsuperscript{\rm 1},
    Zhihu Wang\textsuperscript{\rm 2},\\
    Xu Caiyue\textsuperscript{\rm 2},
    Shiwan Zhao\textsuperscript{\rm 1}\corresponding,
    Qicheng Li\textsuperscript{\rm 1}\corresponding
}
\affiliations{
    \textsuperscript{\rm 1}TMCC, College of Computer Science, Nankai University, Tianjin, China\\
    \textsuperscript{\rm 2}Huawei Technologies Ltd., Beijing, China\\
    xychao@mail.nankai.edu.cn,
    liqicheng@nankai.edu.cn,
    zhaosw@gmail.com
}

\begin{document}
\maketitle

\begin{abstract}
On-policy self-distillation (OPSD) improves reasoning by using a privileged view of a model conditioned on reference solutions to supervise a student view that observes only the question. However, the teacher-provided token-level targets may depend on reference-specific information unavailable at inference time. We propose Problem-Space-Guided OPSD (PS-OPSD), which replaces the complete solution with trajectory-grounded guidance describing the initial state, goal conditions, constraints, and a selected state-transition path. The student rollout and OPSD objective remain unchanged. Across three mathematical reasoning benchmarks and model scales ranging from 1.7B to 8B, PS-OPSD achieves the highest aggregate question-only accuracy among the compared methods. Controlled experiments further indicate that guidance relevance and path coherence contribute to these gains, highlighting the representation of privileged information as an important design choice in OPSD.
\end{abstract}

\section{Introduction}
On-policy self-distillation offers an appealing approach to dense supervision for reasoning.
A student generates its own trajectory, while a privileged view of the same model, conditioned on a verified reference solution, provides next-token supervision along the student-generated prefixes \citep{agarwal2024onpolicy,zhao2026selfdistilled}.
Supervision is therefore applied at states the student actually visits, reducing the mismatch between training prefixes and those encountered during question-only inference.
Because the privileged teacher is another view of the same model, this framework also avoids reliance on a separate, permanently stronger external teacher.

However, its training and deployment interfaces are asymmetric.
During training, the privileged teacher has access to a complete reference solution, whereas the deployed student must reason from the question alone.
A complete solution can make the teacher highly informative, but informativeness does not necessarily imply transferability.
The reference solution ties reusable mathematical relations to a particular final answer, reasoning order, wording, and set of instance-specific calculations.
Consequently, the teacher's token-level targets may depend on information that the student cannot recover from its question-only input at inference time.

Recent analyses of on-policy and privileged self-distillation reinforce this concern.
Teacher choice, information asymmetry, and token-level teachability can substantially affect what is transferred to a question-only student \citep{zhu2026manyfaces,kaur2026rethinking,kim2026degrade,shen2026purified}.
Recent work also reports a concrete behavioral symptom: models trained with OPSD may explicitly refer to reference solutions, answer keys, or hints that are unavailable in their question-only input \citep{yang2026selfdistilledrlvr}.
Such behavior does not by itself establish all forms of privileged-information dependence, but it suggests that conditioning the teacher on a complete reference may introduce reference-specific dependencies that do not transfer cleanly to question-only inference.
This raises a representation question that precedes the design of the distillation loss:
\begin{quote}
\centering
\emph{What privileged information should the teacher observe?}
\end{quote}

We address this question through the classical problem-space view of problem solving, which characterizes a task in terms of an initial state, goal conditions, applicable operators, and constraints on valid state transitions \citep{newell1972human,fikes1971strips}.
Theories of relational transfer further suggest that structural relations are more reusable than the surface form through which a particular solution instantiates them \citep{gentner1983structure,gick1983schema}.
Together, these perspectives motivate representing privileged information not as a complete reference trace, but as the problem-solving structure realized along a verified solution trajectory.

We do not attempt to reconstruct the full problem space, which would require enumerating all reachable states, admissible operators, and alternative solution paths.
Instead, for each problem--solution pair, we extract the portion instantiated by the verified reference trajectory.
The resulting representation, which we call Problem-Space Guidance, records the Initial State, Goal Conditions, Constraints, and a Selected State-Transition Path.
Each transition specifies an Operator, its Preconditions, the induced Transformation, and the Resulting State.
This representation remains grounded in a valid solution while emphasizing functional state changes over trace-specific wording, ordering, and arithmetic details.

\begin{figure*}[t]
    \centering
    \includegraphics[width=\textwidth]{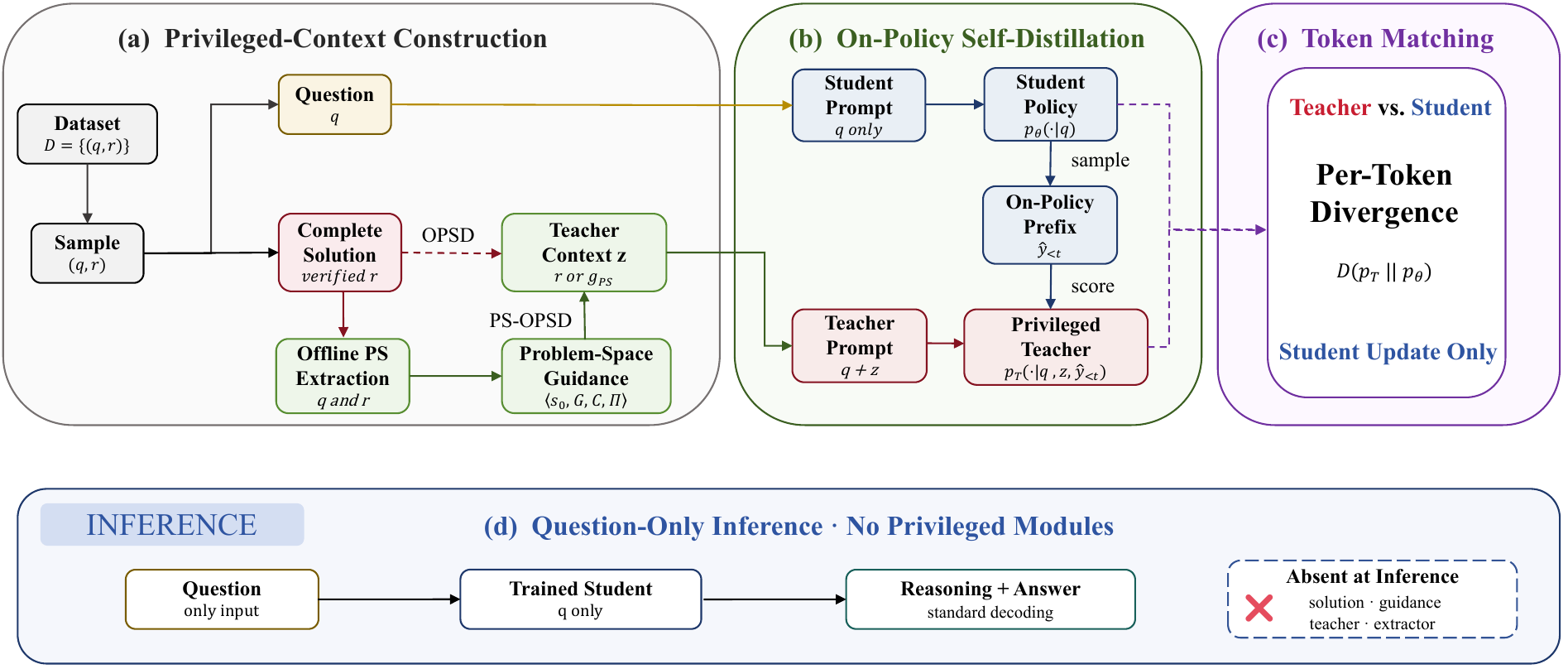}
\caption{PS-OPSD replaces a complete-solution teacher context with explicit, trajectory-grounded problem-space guidance.
OPSD conditions the privileged teacher on the full reference solution, whereas PS-OPSD extracts an initial state, goal conditions, constraints, and a selected state-transition path whose steps specify operators, preconditions, transformations, and resulting states.
In both settings, the question-only student generates on-policy prefixes and receives the same token-level OPSD distribution-matching objective.
At inference time, the trained student observes only the question; the reference solution, guidance, teacher, and extractor are absent.}
\label{fig:ps-opsd-overview}
\end{figure*}

Based on this representation, we propose \textbf{Problem-Space-Guided On-Policy Self-Distillation (\psopsd{})}.
An offline extractor first transforms each verified reference solution into Problem-Space Guidance.
Within the OPSD training loop, \psopsd{} modifies only the privileged teacher context: the student continues to generate question-only on-policy trajectories, and the token-level distillation objective remains unchanged.
At inference time, the trained student reasons from the question alone, without access to the reference solution, extracted guidance, privileged teacher, or extractor.
Figure~\ref{fig:ps-opsd-overview} summarizes the training and inference interfaces.

We evaluate \psopsd{} using Qwen3 models at three scales, from 1.7B to 8B parameters, on three competition-level mathematical reasoning benchmarks.
Across all three scales, \psopsd{} achieves the highest aggregate question-only accuracy among the compared methods.
Controlled representation variants further investigate the roles of guidance relevance, explicit field structure, and state-transition coherence.
We additionally analyze teacher-input exposure and downstream question-only behavior to better understand how the representation of privileged information affects self-distillation.

Our main contributions are:
\begin{itemize}
    \item We formulate privileged-context design in same-model on-policy self-distillation as a transfer problem and introduce trajectory-grounded Problem-Space Guidance.
    \item We develop \psopsd{}, which replaces complete reference solutions in the privileged teacher context with structured guidance while leaving the student rollout and token-level distillation objective unchanged.
    \item We evaluate \psopsd{} across three model scales and three mathematical reasoning benchmarks, and use representation controls and behavioral diagnostics to examine which properties of privileged guidance contribute to its effectiveness.
\end{itemize}
\section{Related Work}
\paragraph{On-policy distillation and self-distillation.}
Knowledge distillation transfers teacher predictions through soft targets or teacher-generated sequences \citep{hinton2015distilling,kim2016sequence}.
When those targets are evaluated only on fixed trajectories, training prefixes can diverge from states visited by the student at inference time, a general sequential-learning problem studied by scheduled sampling and learner-state imitation \citep{bengio2015scheduled,ross2011reduction}.
Generalized Knowledge Distillation moves supervision onto student-generated sequences and supports forward- or reverse-KL targets \citep{agarwal2024onpolicy}.
For mathematical reasoning, GRPO instead supplies sequence-level verifiable rewards on sampled trajectories \citep{shao2024deepseekmath}.
Process reward models provide denser feedback over intermediate reasoning steps, but require separate step-level supervision or reward modeling \citep{lightman2024verify}.
OPSD combines dense on-policy supervision with a same-model teacher: the student generates from the question, while a privileged teacher reads a verified solution and scores the student's prefixes \citep{zhao2026selfdistilled}.
Our work keeps this OPSD mechanism fixed and studies the representation of the privileged context.

\paragraph{Privileged information and transferability.}
Training-only privileged information can improve a deployment-time predictor, and its connection to distillation has long been recognized \citep{vapnik2009lupi,lopezpaz2016unifying}; its utility nevertheless depends on whether the learner can internalize what the privileged view reveals \citep{penaloza2026privileged}.
Recent analyses of on-policy self-distillation sharpen this concern for reasoning models.
Instance-specific context, incompatible teacher behavior, and locally unteachable token preferences can weaken or reverse transfer \citep{zhu2026manyfaces,kaur2026rethinking,wang2026teachability}.
Other work reports that rich solution context may suppress useful epistemic behavior or inject reference-specific shortcut signal \citep{kim2026degrade,shen2026purified}.
AVSD addresses view dependence by combining full-solution, partial-solution, and answer-only teachers into consensus and gated residual signals \citep{nguyen2026avsd}.
In contrast, we ask whether a single teacher view can be redesigned around problem-solving relations before distillation.

\paragraph{Solution-derived plans and scaffolds.}
Several methods transform, filter, or operationalize solution traces.
Studies of chain-of-thought distillation similarly find that rationale organization and compression affect transferred reasoning behavior \citep{chen2025factors,luo2025deconstructing}.
PAINT adapts how much of a solution is exposed and interpolates privileged energy at selected token positions \citep{tan2026paint}.
Adaptive Teacher Exposure learns a controller that selects how much of a reference prefix to reveal during self-distillation \citep{han2026adaptive}.
OPHSD uses a dynamic plan--solve harness to generate assisted behavior and distills that behavior into an unassisted model \citep{zhao2026ophsd}.
SORT uses a reference-derived plan to weight tokens in off-policy repair for all-wrong GRPO groups, while HiLL learns failure-conditioned hints that alter the reasoner's online RL input \citep{le2026sort,xia2026hill}.
Outside unchanged OPSD, SuperCorrect extracts hierarchical thought templates for critique and correction, while Reasoning Scaffolding predicts semantic reasoning signals alongside solution steps \citep{yang2024supercorrect,wen2025scaffolding}.
These methods establish that solution abstraction and procedural scaffolding are not new by themselves.
Our narrower contribution is a fixed offline state-transition representation, rather than an adaptive reference-prefix exposure policy, used only as teacher context with the student rollout and OPSD objective held unchanged.

\section{Problem-Space-Guided OPSD}
\label{sec:method}

\subsection{OPSD Preliminaries}
Let $q$ be a problem and $r$ a verified reference solution.
A question-only student samples a reasoning trajectory
\begin{equation}
    \hat{y} \sim \student(\cdot \mid q).
    \label{eq:student-rollout}
\end{equation}
At token $t$, OPSD queries a privileged view of the same model on the student-generated prefix $\hat{y}_{<t}$.
OPSD conditions that view on the complete solution,
$\teacher(\cdot \mid q,r,\hat{y}_{<t})$, and matches the student distribution to the resulting token-level target \citep{zhao2026selfdistilled}.
The student therefore remains on-policy even though the teacher receives information unavailable at inference time.

\subsection{Explicit Problem-Space Guidance}
Following classical problem-space accounts \citep{newell1972human,fikes1971strips}, we represent a mathematical problem $q$ conceptually as
\begin{equation}
    \problemspace(q)=\langle S,s_0,O,G,C\rangle.
    \label{eq:problem-space}
\end{equation}
Here, $S$ is the state space, $s_0$ the Initial State, $O$ the admissible operators, and $G$ the Goal Conditions; $C$ is a mathematics-specific extension for domain restrictions, global invariants, and validity conditions.

Equation~\ref{eq:problem-space} is conceptual and is not constructed by PS-OPSD.
Because enumerating $S$ and $O$ is generally impractical and one reference solution $r$ reveals only a single realized trajectory, an offline extractor serializes the teacher guidance as
\begin{equation}
    \psguide(q,r)=\langle s_0,G,C,\pi\rangle.
    \label{eq:ps-guidance}
\end{equation}

\begin{figure}[t]
    \centering
    \includegraphics[width=\columnwidth]{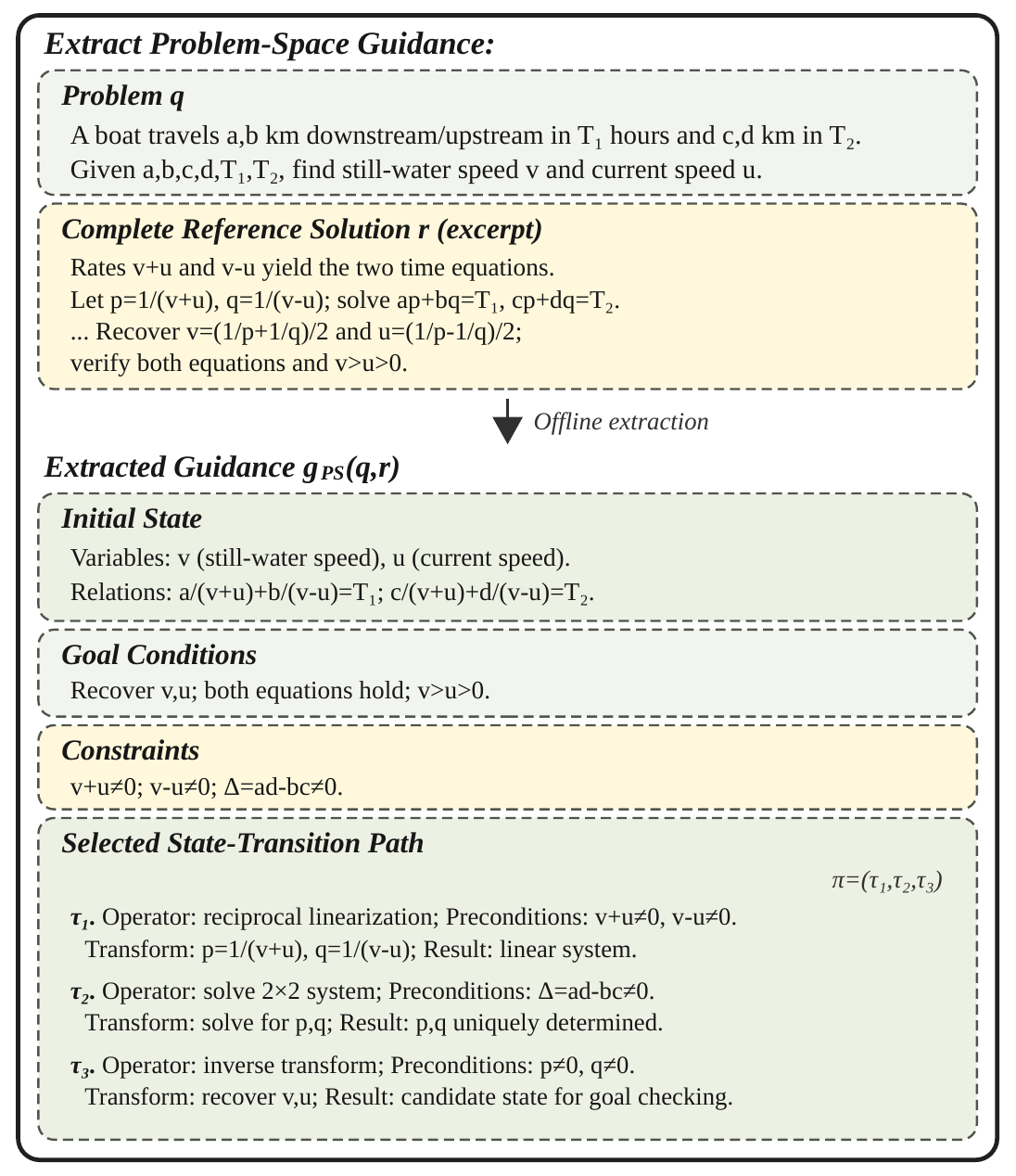}
    \caption{Offline Problem-Space Guidance extraction for a symbolic rate-composition problem.
    The complete verified solution is abbreviated; the four-field output records global problem information and a selected transition path while omitting an instantiated final answer.}
    \label{fig:ps-guidance-example}
\end{figure}

\begin{algorithm}[t]
\hrule
\caption{PS-OPSD Procedure}
\label{alg:ps-opsd}
\hrule
\small
\renewcommand{\arraystretch}{1.08}
\begin{tabular}{@{}r@{\hspace{0.55em}}p{\dimexpr\columnwidth-2em\relax}@{}}
& \textbf{Input:} Dataset $\mathcal{D}=\{(q_i,r_i)\}_{i=1}^{N}$; fixed teacher $p_T$; student $p_\theta$; temperature $T$; clipping threshold $c$.\\
\addlinespace[2pt]
1 & \textbf{Offline extraction:} Cache $g_i\gets\psguide(q_i,r_i)$ for every $(q_i,r_i)\in\mathcal{D}$.\\
\addlinespace[2pt]
2 & \textbf{repeat}\\
3 & \quad Sample a minibatch $\mathcal{B}\subset\{(q_i,g_i)\}_{i=1}^{N}$.\\
4 & \quad \textbf{for each} $(q,g)\in\mathcal{B}$ \textbf{do}\\
5 & \qquad Sample an on-policy trajectory $\hat{y}\sim p_\theta(\cdot\mid q)$.\\
6 & \qquad \textbf{for} $t=1,\ldots,|\hat{y}|$ \textbf{do}\\
7 & \qquad\quad Evaluate $p_{T,t}^{\mathrm{PS}}=p_T(\cdot\mid q,g,\hat{y}_{<t})$.\\
8 & \qquad\quad Evaluate $p_{\theta,t}=p_\theta(\cdot\mid q,\hat{y}_{<t})$.\\
9 & \qquad \textbf{end for}\\
10 & \qquad Compute the mean clipped-KL loss $\ell(q,g)$ over the trajectory.\\
11 & \quad \textbf{end for}\\
12 & \quad Update only $\theta$ using the minibatch mean loss; keep $p_T$ and the cached guidance fixed.\\
13 & \textbf{until} the training budget is exhausted.\\
\addlinespace[2pt]
& \textbf{Output:} Trained question-only student $p_\theta(\cdot\mid q)$.\\
\end{tabular}
\par\hrule
\end{algorithm}

Leaving $S$ and $O$ implicit, the selected path $\pi=(\tau_1,\ldots,\tau_k)$ contains only transition records derived from $r$.
Each transition is serialized as
\begin{equation}
    \tau_i=(o_i,\mathrm{pre}_i,\mathrm{transform}_i,s_i).
    \label{eq:transition-record}
\end{equation}
Given the top-level Initial State $s_0$, each $\tau_i$ records one subsequent transition: $o_i$ names the Operator selected from the conceptual set $O$, $\mathrm{pre}_i$ states its local Preconditions in $s_{i-1}$, $\mathrm{transform}_i$ describes the induced change, and $s_i$ records the Resulting State.
Under the intended semantics, the states $s_0,\ldots,s_k$ respect the global Constraints $C$, and the terminal state $s_k$ meets the Goal Conditions $G$.

Accordingly, Problem-Space Guidance uses four top-level fields:
\begin{quote}
\small
\textbf{Initial State:} variables, entities, givens, and relations.\\
\textbf{Goal Conditions:} target and verifiable success conditions.\\
\textbf{Constraints:} invariants, domains, and validity conditions.\\
\textbf{Selected State-Transition Path:} an ordered list of Operator, Preconditions, Transformation, and Resulting State records.
\end{quote}

Figure~\ref{fig:ps-guidance-example} illustrates how a question and complete verified solution are transformed into this serialization for a symbolic rate-composition problem.
The first three fields specify global problem information, while the final field records a reference-derived sequence of local transitions.

Algorithm~\ref{alg:ps-opsd} summarizes the complete procedure.
The extraction stage is performed once before training; the cached guidance is used only by the teacher, which scores the same prefixes generated and evaluated by the student.


\subsection{Training and Question-Only Inference}
PS-OPSD replaces $r$ in the privileged teacher context with $\psguide(q,r)$ while leaving the question-only student unchanged:
\begin{equation}
    \begin{aligned}
        p_{T,t}^{\mathrm{OPSD}}
        &= \teacher(\cdot \mid q,r,\hat{y}_{<t}), \\
        p_{T,t}^{\mathrm{PS}}
        &= \teacher(\cdot \mid q,\psguide(q,r),\hat{y}_{<t}), \\
        p_{\theta,t}
        &= \student(\cdot \mid q,\hat{y}_{<t}).
    \end{aligned}
    \label{eq:ps-teacher}
\end{equation}
We treat $p_T$ as a fixed target distribution and optimize only the student parameters $\theta$.
For distributions with logits $z_p$ and $z_q$, define the implemented coordinate-clipped KL-form objective as
\begin{equation}
    \begin{aligned}
        p^{(T)} &= \operatorname{softmax}(z_p/T), \\
        q^{(T)} &= \operatorname{softmax}(z_q/T), \\
        d_T(p,q;v)
        &= p^{(T)}(v)\log\frac{p^{(T)}(v)}{q^{(T)}(v)}, \\
        \widetilde{D}_{c,T}(p\|q)
        &= \sum_{v\in\mathcal{V}}\min\!\left\{d_T(p,q;v),c\right\}.
    \end{aligned}
    \label{eq:clipped-kl}
\end{equation}
For the training distribution $\mathcal{D}$, the objective is then
\begin{equation}
\begin{aligned}
\mathcal{L}_{\mathrm{PS\text{-}OPSD}}
&=
\mathbb{E}_{\substack{(q,r)\sim\mathcal{D}\\
\hat{y}\sim\student(\cdot\mid q)}}
\left[
\frac{1}{|\hat{y}|}\sum_t
\widetilde{D}_{c,T}\!\left(p_{T,t}^{\mathrm{PS}}\,\|\,p_{\theta,t}\right)
\right].
\end{aligned}
\label{eq:ps-opsd-objective}
\end{equation}
The vocabulary sum is computed in full, and clipping is applied to each vocabulary-coordinate contribution before summation and averaging over non-padding sequence positions.
Thus, $\widetilde{D}_{c,T}$ is not the unmodified mathematical KL divergence.
The exact temperature and clipping threshold remain bound to each run manifest because the clipping configuration differs across scales and runs.

No guidance is appended to the student prompt.
At inference time, generation follows $\student(\cdot\mid q)$ without the reference, extractor, teacher, or any planning module.

\section{Experimental Setup}
\label{sec:setup}

\paragraph{Models and training data.}
We evaluate Qwen3-1.7B, Qwen3-4B, and Qwen3-8B \citep{yang2025qwen3}.
Teacher and student use the same model family and initialization within each condition.
In all OPSD variants, the teacher is the initial Qwen3 checkpoint evaluated with LoRA adapters disabled and without gradient tracking, while optimization updates only the student's LoRA parameters.
Training uses the mathematical-reasoning subset of OpenThoughts \citep{guha2025openthoughts}.

\paragraph{Guidance extractor.}
We use Qwen3.6-35B-A3B \citep{qwen2026qwen36} as an offline extractor to transform each question and verified reference solution into Problem-Space Guidance.
The resulting guidance is cached before training, and the extractor is not used during optimization or question-only inference.

\begin{table*}[!t]
\centering
\small
\begin{tabular*}{\textwidth}{@{\extracolsep{\fill}}lcccc@{}}
\toprule
Method & AIME24 & AIME25 & HMMT25 & Avg \\
\midrule
\multicolumn{5}{l}{\textit{Qwen3-1.7B}} \\
Base & 48.89 & 38.33 & 22.78 & 36.67 \\
SFT & $51.11 \pm 0.28$ & $38.70 \pm 0.70$ & $26.76 \pm 0.70$ & $38.86 \pm 0.39$ \\
GRPO & $49.44 \pm 0.56$ & $40.19 \pm 0.32$ & $23.43 \pm 0.98$ & $37.69 \pm 0.09$ \\
\opsd{} & $53.80 \pm 0.89$ & $\mathbf{41.67 \pm 0.83}$ & $27.78 \pm 0.56$ & $41.08 \pm 0.28$ \\
AVSD & $51.85 \pm 0.16$ & $38.61 \pm 0.73$ & $26.57 \pm 1.53$ & $39.01 \pm 0.57$ \\
\psopsd{} & $\mathbf{55.37 \pm 1.53}$ & $41.48 \pm 2.89$ & $\mathbf{32.50 \pm 1.94}$ & $\mathbf{43.12 \pm 2.00}$ \\
\midrule
\multicolumn{5}{l}{\textit{Qwen3-4B}} \\
Base & 75.28 & 66.39 & 43.61 & 61.76 \\
SFT & $74.35 \pm 0.42$ & $66.30 \pm 1.53$ & $44.17 \pm 1.00$ & $61.60 \pm 0.79$ \\
GRPO & $75.74 \pm 0.85$ & $66.57 \pm 1.40$ & $44.81 \pm 0.70$ & $62.38 \pm 0.84$ \\
\opsd{} & $73.06 \pm 0.73$ & $68.06 \pm 1.00$ & $43.98 \pm 0.70$ & $61.70 \pm 0.79$ \\
AVSD & $75.46 \pm 0.70$ & $66.67 \pm 0.28$ & $\mathbf{46.20 \pm 0.42}$ & $62.78 \pm 0.19$ \\
\psopsd{} & $\mathbf{76.57 \pm 0.85}$ & $\mathbf{70.56 \pm 0.73}$ & $45.83 \pm 1.21$ & $\mathbf{64.32 \pm 0.33}$ \\
\midrule
\multicolumn{5}{l}{\textit{Qwen3-8B}} \\
Base & 75.00 & 66.94 & 45.00 & 62.31 \\
SFT & $76.02 \pm 0.58$ & $67.78 \pm 1.69$ & $43.80 \pm 1.63$ & $62.53 \pm 1.02$ \\
GRPO & $75.83 \pm 0.48$ & $69.35 \pm 0.58$ & $44.72 \pm 1.00$ & $63.30 \pm 0.46$ \\
\opsd{} & $75.74 \pm 0.98$ & $70.09 \pm 0.89$ & $44.91 \pm 0.98$ & $63.58 \pm 0.62$ \\
AVSD & $\mathbf{77.78 \pm 0.73}$ & $66.67 \pm 1.00$ & $46.48 \pm 1.25$ & $63.64 \pm 0.96$ \\
\psopsd{} & $77.13 \pm 0.85$ & $\mathbf{71.30 \pm 0.58}$ & $\mathbf{47.78 \pm 0.28}$ & $\mathbf{65.40 \pm 0.21}$ \\
\bottomrule
\end{tabular*}
\caption{Question-only generation accuracy across model scales (\%). All scores use 12 stored generations per problem. For each post-training method and benchmark, the checkpoint with the highest accuracy averaged across three independent training seeds is selected; entries report the mean $\pm$ sample standard deviation across those seeds. Avg is the unweighted mean across the three benchmarks and may therefore combine checkpoints. Bold marks the highest mean in each scale and column.}
\label{tab:main-results}
\end{table*}

\paragraph{Benchmarks and metrics.}
We evaluate competition-level mathematical reasoning on AIME24, AIME25, and HMMT25.
All tabulated accuracy results use $K=12$ stored question-only generations per problem, produced under common decoding settings and generation limits.
For a benchmark with $N$ problems, accuracy is reported as a percentage,
\begin{equation}
    \mathrm{Accuracy}=\frac{100}{KN}\sum_{i=1}^{N}c_i,
    \label{eq:benchmark-accuracy}
\end{equation}
where $K=12$ and $c_i$ is the number of correct generations among the $K$ generations for problem $i$.
For each scale, Avg is the unweighted mean of the three benchmark accuracies.

\paragraph{Compared methods.}
The main comparison includes the unadapted post-trained Qwen3 checkpoint (Base), supervised fine-tuning (SFT), GRPO \citep{shao2024deepseekmath}, OPSD \citep{zhao2026selfdistilled}, AVSD \citep{nguyen2026avsd}, and \psopsd{}.
OPSD exposes the complete verified solution to the privileged teacher.
AVSD combines multiple privileged views and reconstructs a consensus-plus-residual target, so it is a method-level baseline rather than a representation control.

\paragraph{Representation controls.}
At the 4B primary setting, we compare Problem-Space Guidance with three controlled variants.
\textbf{Flattened PS} removes field names, hierarchy, and step delimiters while preserving content, order, and token budget under a deterministic serialization rule.
\textbf{Mismatched PS} replaces the target guidance with a donor from another problem, matched by problem family, difficulty bin, and guidance length.
\textbf{Path-Corrupted PS} retains the semantics inside every transition but breaks global path coherence by reordering or incompatibly composing transitions.
The corruption audit verifies local-content retention and cross-step dependency breakage.

\paragraph{Correct-path token entropy.}
We examine whether the accuracy gains of OPSD and \psopsd{} are merely associated with a sharper predictive distribution along correct responses.
Each method and scale contributes the checkpoint with the highest Avg across the three benchmarks, with ties broken by the earlier step.
We replay each checkpoint's 12 question-only generations under teacher forcing and compute
$H_t=-\sum_v p_\theta(v\mid q,y_{<t})\log p_\theta(v\mid q,y_{<t})$
over the raw full-vocabulary distribution at temperature one.
We average over non-structural response tokens within each path, then micro-average over all, correct, or failed paths.
Confidence intervals resample matched problems with all 12 paths and are stratified by benchmark.

\section{Results and Analysis}
\label{sec:results}

\subsection{Main Question-Only Performance}
Table~\ref{tab:main-results} implements the paper's consistency criterion.
For each scale, we first average accuracy over AIME24, AIME25, and HMMT25; \psopsd{} counts as consistently highest within the oracle envelope only if its Avg strictly exceeds every compared method at all three scales.
The observed oracle-envelope Avg scores are 43.12, 64.32, and 65.40 for \psopsd{} at 1.7B, 4B, and 8B, exceeding the strongest baseline at every scale by 1.54--2.04 percentage points.
Across the nine scale--benchmark cells, \psopsd{} is strictly highest in six.
The three exceptions are OPSD on AIME25 at 1.7B and AVSD on HMMT25 at 4B and AIME24 at 8B, where the respective gaps over \psopsd{} are 0.19, 0.37, and 0.65 points; the headline result is therefore cross-benchmark consistency, not uniform per-benchmark dominance.
We report every benchmark even when its direction differs from the aggregate result, and we treat exact ties as co-highest rather than uniquely highest.

\begin{figure*}[!t]
    \centering
    \includegraphics[width=\textwidth]{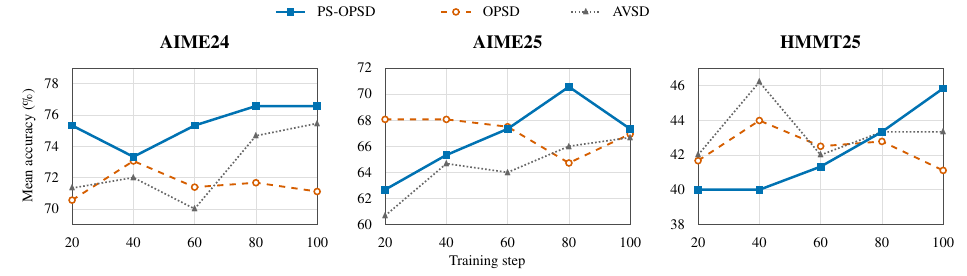}
    \caption{Question-only accuracy (\%) across checkpoints 20--100 for OPSD, AVSD, and PS-OPSD at Qwen3-4B.
    The panels report AIME24, AIME25, and HMMT25, respectively.}
    \label{fig:training-dynamics-4b}
\end{figure*}

Figure~\ref{fig:training-dynamics-4b} complements the oracle envelope with matched-checkpoint trajectories at the 4B primary setting.
At checkpoint 100, \psopsd{} exceeds \opsd{} on all three benchmarks.
It also has the highest endpoint accuracy among the three methods on all three benchmarks.
The trajectories also show that the endpoint advantage emerges late rather than holding uniformly throughout training.

\begin{figure*}[t]
    \centering
    \includegraphics[width=0.78\textwidth]{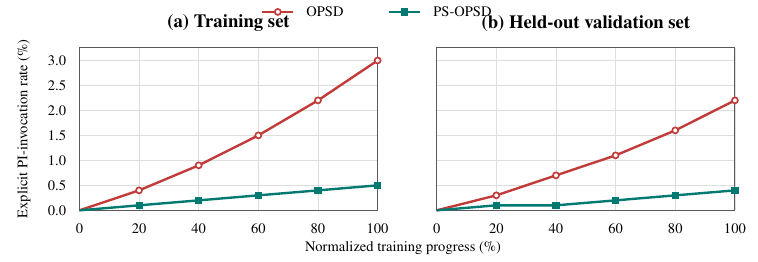}
    \caption{Measured explicit PI-invocation rates for OPSD and PS-OPSD over normalized training progress on (a) training problems and (b) held-out validation problems.}
    \label{fig:pi-invocation}
\end{figure*}

Following \citet{yang2026selfdistilledrlvr}, who characterize explicit appeals to an invisible reference solution at inference time as privileged information leakage, Figure~\ref{fig:pi-invocation} measures this behavioral manifestation.
A completion is positive if it explicitly invokes an unavailable reference solution, answer key, hint, or guidance.
On the training problems, the invocation rate at the final checkpoint is 3.0\% for \opsd{} and 0.5\% for \psopsd{}; on held-out validation problems, it is 2.2\% and 0.4\%, respectively.
The lower rates for \psopsd{} on both splits are consistent with reduced reliance on unavailable reference-specific information, including beyond the problems encountered during optimization.

\subsection{What Matters in Problem-Space Guidance}
\nopagebreak[4]
\par\medskip
\noindent\begin{minipage}{\columnwidth}
\centering
\small
\setlength{\tabcolsep}{3.5pt}
\begin{tabular}{@{}lcccc@{}}
\toprule
Condition & AIME24 & AIME25 & HMMT25 & Avg \\
\midrule
\psopsd{} & \textbf{76.57} & \textbf{70.56} & 45.83 & \textbf{64.32} \\
Flattened PS & 75.37 & 68.24 & \textbf{47.22} & 63.61 \\
Mismatched PS & 74.63 & 67.31 & 43.33 & 61.76 \\
Path-Corrupted PS & 73.06 & 68.89 & 44.63 & 62.19 \\
\bottomrule
\end{tabular}
\captionof{table}{Controlled guidance comparison at Qwen3-4B (accuracy, \%). Each score is the per-benchmark oracle-best checkpoint selected on the same benchmark being reported. Avg is the unweighted mean of the three displayed scores. Bold marks the highest value in each column.}
\label{tab:controls}
\end{minipage}
\par\medskip
Table~\ref{tab:controls} probes three representation hypotheses at the 4B primary setting.
\psopsd{} reaches an Avg of 64.32, compared with 63.61 for Flattened PS, 61.76 for Mismatched PS, and 62.19 for Path-Corrupted PS.
The 0.71-point Avg gap over Flattened PS, computed from unrounded values, is consistent with an incremental role for visible fields and boundaries in aggregate, but Flattened PS is 1.39 points higher on HMMT25, so the pattern is not uniform across benchmarks.
The 2.56-point gap over Mismatched PS appears on all three benchmarks and is consistent with the importance of target-relevant guidance.
The 2.13-point gap over Path-Corrupted PS also appears on all three benchmarks and is consistent with sensitivity to globally coherent transition order.

\subsection{Problem-Level Behavior}
Aggregate accuracy does not distinguish general post-training gains from the
incremental effect of changing the privileged representation.  Comparisons
with Base measure the overall effect of post-training.  \opsd{} and \psopsd{}
use the same OPSD objective and differ only in the privileged teacher context:
\opsd{} receives the complete reference solution, whereas \psopsd{} receives
Problem-Space Guidance.  Their comparison therefore focuses on the choice of
privileged representation.

\par\medskip
\noindent\begin{minipage}{\columnwidth}
\centering
\small
\setlength{\tabcolsep}{4.5pt}
\begin{tabular}{@{}lrrrr@{}}
\toprule
Method & $\Delta$Base & Better & Worse & Tied \\
\midrule
Base & $+0.00$ & 0 & 0 & 90 \\
SFT & $-0.15$ & 13 & 15 & 62 \\
GRPO & $+0.62$ & 17 & 13 & 60 \\
\opsd{} & $-0.06$ & 13 & 11 & 66 \\
AVSD & $+1.02$ & 13 & 11 & 66 \\
\psopsd{} & $+2.56$ & 18 & 7 & 65 \\
\bottomrule
\end{tabular}
\captionof{table}{Problem-level capability changes relative to Base for Qwen3-4B over 90 benchmark problems. $\Delta$Base is the aggregate accuracy difference in percentage points. Better, Worse, and Tied count problems on which a method produces more, fewer, or the same number of correct generations as Base.}
\label{tab:problem-behavior}
\end{minipage}
\par\medskip
Table~\ref{tab:problem-behavior} first asks whether each training method
improves over the unadapted model.  \psopsd{} has the largest aggregate gain
at $+2.56$ points, improving the per-problem correct count on 18 problems and
regressing on 7.  AVSD and GRPO improve by $+1.02$ and $+0.62$ points,
respectively, while \opsd{} and SFT decline by $0.06$ and $0.15$ points.

The direct comparison between \psopsd{} and \opsd{} asks a narrower
representation question.  \psopsd{} has a $+2.62$-point aggregate difference,
with 23 problems favoring \psopsd{}, 9 favoring \opsd{}, and 58 tied.

\noindent\begin{minipage}{\columnwidth}
\centering
\small
\setlength{\tabcolsep}{3.5pt}
\begin{tabular}{@{}lrrrr@{}}
\toprule
Method & Hard Avg & Hard Pass & Ret. & Net Exp. \\
\midrule
Base & 0.00 & 0.00 & 100.00 & 0 \\
SFT & 0.85 & 4.35 & 97.01 & $-1$ \\
GRPO & 6.04 & 13.04 & 98.51 & $+2$ \\
\opsd{} & 1.69 & 8.70 & 97.01 & 0 \\
AVSD & 2.66 & 4.35 & 98.51 & 0 \\
\psopsd{} & 4.35 & 13.04 & 98.51 & $+2$ \\
\bottomrule
\end{tabular}
\captionof{table}{Base-defined hard-case rescue at Qwen3-4B (\%).}
\label{tab:hard-case-rescue}
\end{minipage}
Table~\ref{tab:hard-case-rescue} separates new capability from retention of
problems that Base already solves.  Relative to \opsd{}, \psopsd{} raises
hard-case accuracy from $1.69$ to $4.35$, hard-case pass rate from $8.70\%$ to
$13.04\%$, and retention from $97.01\%$ to $98.51\%$.  Its net expansion is
$+2$, compared with 0 for \opsd{}, showing that its aggregate gain is
accompanied by broader coverage of the Base-hard subset while retaining
$98.51\%$ of Base-solved problems.

\subsection{Correct-Path Token Entropy}
\nopagebreak[4]
\par\medskip
\noindent\begin{minipage}{\columnwidth}
\centering
\small
\begin{tabular}{@{}llrrr@{}}
\toprule
Scale & Reference & $\Delta$Avg & $\Delta$pass@12 & $\Delta H_C$ \\
\midrule
1.7B & Base & $+5.34$ & $-1.11$ & $+0.0697$ \\
1.7B & \opsd{} & $-0.68$ & $-4.44$ & $-0.0185$ \\
4B & Base & $+0.43$ & $+0.00$ & $+0.0763$ \\
4B & \opsd{} & $+0.00$ & $+1.11$ & $+0.0321$ \\
8B & Base & $+1.11$ & $+2.22$ & $+0.0453$ \\
8B & \opsd{} & $+0.68$ & $+4.44$ & $+0.0319$ \\
\bottomrule
\end{tabular}
\captionof{table}{Paired best-checkpoint differences, \psopsd{} minus the reference. Accuracy columns are percentage points; $H_C$ is correct-path token entropy in nats. Positive $\Delta H_C$ indicates a less concentrated token distribution along each method's own correct paths, not greater semantic strategy diversity or better reasoning by itself.}
\label{tab:path-entropy}
\end{minipage}
\par\medskip
Table~\ref{tab:path-entropy} uses Base to calibrate how OPSD training changes predictive concentration.
At the selected checkpoints, both \opsd{} and \psopsd{} improve Avg over Base at all three scales, and neither exhibits lower correct-path token entropy.
The implied \opsd{}--Base entropy differences are $+0.0882$, $+0.0442$, and $+0.0134$ at 1.7B, 4B, and 8B, respectively.
Thus, the observed Avg gains are not obtained through a uniformly sharper correct-path distribution; pass@12 relative to Base remains mixed.

The direct comparison between \psopsd{} and \opsd{} is capacity dependent.
At 1.7B, \psopsd{} has lower Avg, pass@12, and correct-path entropy.
At 4B, Avg is unchanged, while pass@12 and entropy increase by $1.11$ points and $0.0321$ nats.
At 8B, Avg, pass@12, and entropy increase by $0.68$, $4.44$, and $0.0319$, respectively.
The 4B and 8B results show that \psopsd{}'s coverage gains do not rely on a more concentrated predictive distribution.

\section{Conclusion and Limitations}
\psopsd{} achieves the highest Avg among the compared methods at every model scale, exceeding the strongest baseline by 1.54--2.04 percentage points.
The representation controls further support the importance of target-relevant guidance and globally coherent transition paths, while the explicit PI-invocation rate is lower than OPSD on both training and held-out problems.
Together, these results show that more privileged information is not automatically more transferable: structuring the teacher context around problem-solving relations provides a stronger interface for question-only self-distillation than exposing the complete reference solution.

Our evaluation is limited to competition mathematics, three Qwen3 scales.
Problem-Space Guidance captures a selected trajectory derived from one verified solution rather than the complete problem space, and the offline extractor adds preprocessing cost.
Future work should test other model families and domains, alternative verified paths, recovery from intermediate errors.
The teacher-input exposure and PI-invocation analyses are surface and behavioral diagnostics rather than causal identification of privileged information leakage, while correct-path token entropy measures local predictive concentration rather than semantic strategy diversity.
\clearpage
\bibliography{references}
\end{document}